\documentclass{article}

\usepackage{arxiv}

\usepackage[utf8]{inputenc} 
\usepackage[T1]{fontenc}    
\usepackage{hyperref}       
\usepackage{url}            
\usepackage{booktabs}       
\usepackage{amsfonts}       
\usepackage{nicefrac}       
\usepackage{microtype}      
\usepackage{lipsum}
\usepackage{graphicx}
\usepackage{float}
\graphicspath{ {./images/} }

\title{Time out of Mind: Generating Rate of Speech conditioned on emotion and speaker}
\author{%
  Navjot Kaur \\
  \texttt{nka77@sfu.ca} \\
  \And
  Paige Tuttosi\\
  \texttt{ptuttosi@sfu.ca} \\
}

\begin{document}
\maketitle
\begin{abstract}
  Voice synthesis has seen significant improvements in the past decade resulting in highly intelligible voices. Further investigations have resulted in models that can produce variable speech, including conditional emotional expression. The problem lies, however, in a focus on phrase-level modifications and prosodic vocal features. Using the CREMA-D dataset we have trained a GAN conditioned on emotion to generate worth lengths for a given input text. These word lengths are relative to neutral speech and can be provided, through speech synthesis markup language (SSML) to a  text-to-speech (TTS) system to generate more expressive speech. Additionally, a generative model is also trained using implicit maximum likelihood estimation (IMLE) and a comparative analysis with GANs is included. We were able to achieve better performances on objective measures for neutral speech, and better time alignment for happy speech when compared to an out-of-box model. However, further investigation of subjective evaluation is required. 
\end{abstract}


\section{Introduction}
As humans, we are particularly fascinated by the aspects of ourselves that are difficult to put words to, yet are inherent to our intrinsic humanness. We want to be able to define the nature of these fundamental human instincts and, as an extension, be able to replicate them. One of these human aspects of particular interest is emotionality and expressivity. Researchers have been distinctly interested in being able to generate naturalistic human emotions. Yet, the interpretation and generation of image-based features, e.g. creating faces, expressions, and gestures, has seen significantly more research capacity than vocal features, despite the importance of voice  as a social cue \cite{maruri21_interspeech}.

For many years, the primary concern for voice generation was intelligibility. Because of this, state-of-the-art generated voices can be mistaken for a human voice, but they still lack contextual adaptation, specifically for non-prosodic features. Features such as pause rate, word length, and spectral features have been poorly explored in vocal synthesis despite their impact on vocal perception \cite{ours}. This is in part due to the limitations of SSML and the nature of these features belonging to the time, rather than the frequency domain. This requires sequential generation techniques. A second issue is most systems focus on phrase level modifications\cite{CAMNet}. In some cases, random variations are incorporated over the phrase given the variance of the feature \cite{jobs}. However, little attention has been given to word-level manipulation, despite indications that these fine-grained modifications are of particular importance in human speech \cite{speakeridentity}.

We propose a generative model to take text as an input, and when conditioned on an emotion, will produce a set of word lengths for this phrase. This model is 1. Linguistically contextual: Linguistic features of the input text are considered when generating word lengths. 2. Granular: The result is per word rather than an average over a phrase. 3. Prosody independent: Input prosody is not required to generate word lengths, only text. 4. Sequential: The sequential context of a text is considered.

\section{Related Works}

\subsection{Emotional Speech}
Most often human vocal modifications are for the purposes of creating ‘deliberately clear speech,' \cite{confluent_talker_and_listener}. In some cases, speech is not modified for clarity, but rather to communicate a specific emotion of purpose, such as politeness\cite{polite}. Vocal modifications are produced without conscious effort to elicit a specific auditory feature, rather they are produced as a result of achieving the aforementioned goals. A table of emotional vocal modifications for the basic emotions can be seen in Appendix \ref{Figures} Fig.\ref{emotion-table}.

\subsection{Speech Synthesis}
TTS has become an inexpensive and efficient means to create realistic voices  \cite{TTS_relevancy_of_voice_effect, cost_effective_cvocoder, TTS_and_robots}. Companies like Google\footnote{https://cloud.google.com/text-to-speech/}, Amazon\footnote{https://aws.amazon.com/polly/}, and Microsoft\footnote{https://azure.microsoft.com/en-us/services/cognitive-services/text-to-speech/} all have their own variations of these vocoders. Tacotron-GST even expresses basic emotions, however, the modifications are only in the frequency domain, i.e. prosodic \cite{predict_expressive_speaking,Expressive_tts, generating_diverse_tts, CAMNet}. Furthermore, TTS is constrained by SSML. Although the available features have broadened and include loudness, pitch, and rate-of-speech\footnote{https://cloud.google.com/text-to-speech/docs/ssml} these must be manually manipulated; as such there is a need to generate values for these features. 

\subsection{Emotion-aware content generation}
The recent works on generating content conditioned on emotion use variants of conditional GANs \cite{emotalkingface, hifiGan, hifiGanB}. In this work, we have used Wasserstein GAN for generating relative word lengths, for the first time to our knowledge. We also experiment with the implicit maximum likelihood estimation model \cite{imle} to generate more robust and `realistic' data.

\section{Dataset}\label{Dataset}
We used the CREMA-D dataset \cite{HouweiCao2014CCEM}. This dataset consists of 91 actors, 48 males and 43 females between the ages of 20 and 74. The actors came from a variety of races and ethnicities (African America, Asian, Caucasian, Hispanic, and Unspecified). Leading up to and during recording an acting coach was present to help induce emotion in the actors. The actors spoke 12 sentences that were determined to be emotionally neutral \cite{RussJeffB.2008Voaa}. The sentences are listed in Section \ref{crema_sentences}.

The actors produced 6 basic emotions : anger, disgust, fear, happiness, neutral and sadness. Surprise, although considered to be one of the basic emotions, was not included as the acting coach suggested this emotion contained too many conflicting sub emotions to be of use. Each of the emotions was produced at one of 4 levels : low, medium, high, or unspecified.

This work produced 7442 clips that are each available as multi-modal video, audio, or image only video. The dataset was then validated by 2443 raters. Each validater rated 90 unique clips, 30 audio, 30 visual, and 30 audio-visual. This resulted in 95\% of the clips having more than 7 ratings. Of these validations the intended emotion was selected 40.9\%, 58.2\% and 63.6\% of the time for audio-only, visual-only, and audio-visual data respectively.

\subsection{Data Preparation}
To prepare our data we wanted to be sure we were only including audio clips of high quality emotional replication. To ensure this we used the Krippendorff's alpha provided by the author. Krippendorff's alpha is a measure of inter-rater agreement that is able to handle categorical and missing responses \cite{K.Klaus1970Etrs} \cite{K.Klaus2004Rica}. An $\alpha \geq 0.823$ is considered to be a good agreement, with $ 0.667 \leq \alpha \leq 0.823$ considered acceptable agreement \cite{alpha.agree}. We decided to use the cutoff of 0.667 for audio-only rating (acceptable agreement) as this resulted in 3413 audio clips, which already greatly reduced our training sample. We then defined the ground truth label as the final label for the model since, given an acceptable rater agreement, this should be representative of perceived emotion.

To extract linguistic information from the 12 phrases we used the spaCy library\footnote{https://spacy.io/}. We extracted part of speech tagging (POS), dependency paring (DEP) and lemmatization (lemma). We then extracted word lengths and pause lengths with Gentle Aligner \footnote{https://github.com/lowerquality/gentle}. Gentle Aligner maps timestamps on both the word and phoneme level, including the start and end times of each word. Using these start and end times we were able to calculate the duration of each word. To ensure that duration is independent of word length we calculated each of the word lengths relative to the neutral word in the given phrase for each of the emotions. This also results in an optimal encoding for SSML as a rate of speech tag can be provided as a percentage relative to neutral speech, e.g. <prosody rate="50\%">.

\section{Modelling}
\begin{figure}[ht]
      \centering      \includegraphics[width=0.8\linewidth,height=0.25\linewidth]{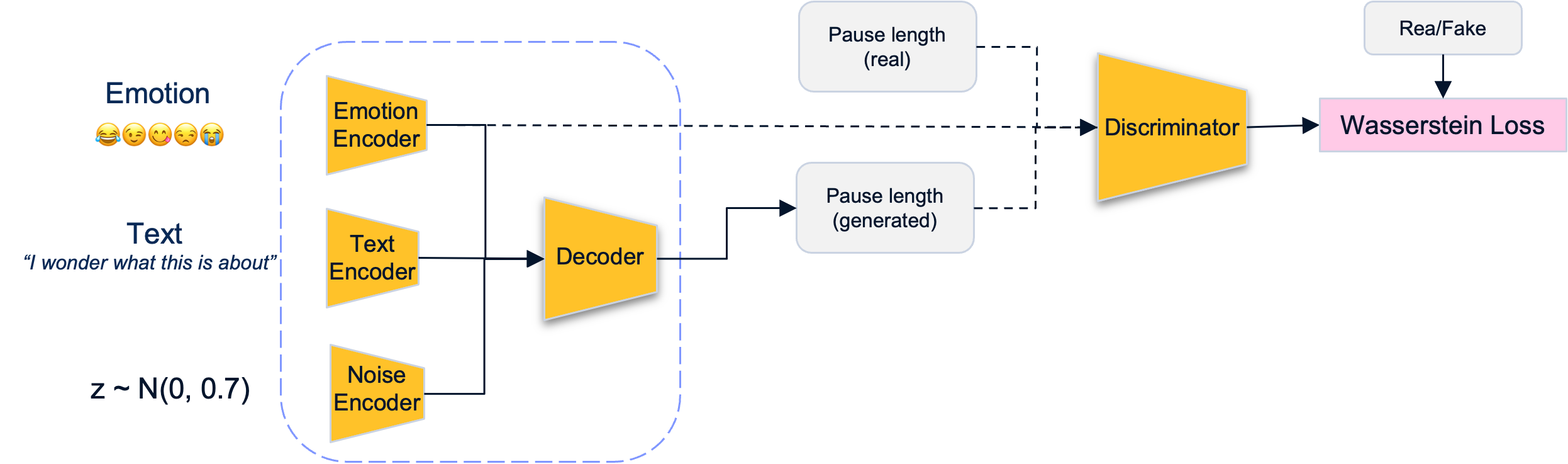}
      \caption{GAN model pipeline}
      \label{GAN-model}
\end{figure}

Given a text sequence, we want to generate word length relative to a neutral speech, for each word in a phrase. We use emotion as an input condition so that the output word length can be manipulated to generate speech of varying emotions, like happy or angry. To model this, we consider a conditional generative model architecture, specifically Wasserstein GANs \cite{WGAN}, as shown in Fig. \ref{GAN-model}. The generator architecture takes in emotion and text, along with random noise, and generates a sequence of word lengths. The discriminator network learns to differentiate between the generated word length against the ground truth. 

\subsection{GAN architecture}
Inspired by \cite{emotalkingface}, the generator network is made up of multiple sub-networks: text, emotion and noise encoders and a word length decoder.

The text encoder takes text as input in form of a sequence of 1-hot encoding vectors and maps it into a text latent code. Given the limited size of the vocabulary in the CREMA dataset the size of the 1-hot vector is 54. These vectors are mapped into dense space using a linear neural layer. The vector sequence is then passed into bi-directional LSTM module to learn the sequential context, followed by another linear layer to further reduce the dimensionality of the latent code. This results in a sequence of latent codes of length 20.
The emotion encoder is a simple stack of two linear layers which take emotion as a 1-hot encoded vector and map it to a latent space of 3 length. All the outputs of the text and emotion are concatenated together.

The noise encoder is also a simple linear layer that takes noise as its input. The noise is sampled from a normal distribution with a mean of 0 and a variance of 0.7. The variance is chosen such that it is similar to the value of the rest of the encoders. The noise is stacked and reshaped such that the dimensions are the same as the outputs from the rest of the encoders and then added to the outputs. This gives the final latent code for the generator.

The decoder takes in the latent code and learns to construct the word length sequence relative to the neutral speech. Since the output is sequential in nature, we use a stack of two LSTM layers followed by a linear layer to learn the desired sequence. This sequence represents the generated word length for speech. In addition, another head with a linear layer is used to predict the emotion from the generated output to ensure that the emotional context is captured correctly.

The discriminator takes in word length sequences and learns to predict whether it belongs to the real or fake (generated) data class. We use two layers of LSTM modules, along with linear layers for learning the classification.
\subsection{Objective function}\label{loss_func}
The discriminator loss is the negative Wasserstein distance between the generator distribution and the real data distribution. It is an approximation of the Earth Mover (EM) distance, which theoretically shows that it can gradually optimize the training of the GAN. The measure is approximated by enforcing the K-Lipschitz constraint on the weights so that they lie in a compact space. One easy approach to enforce this is to clamp the weights between a small region. We consider [-0.03, 0.03] for our experiments. The discriminator $D$ tries to maximize $D(x) - D(G(z))$ while the generator $G$ tries to maximize $D(G(z))$, where $x$ is real and $z$ is generated data. The generator loss is updated less often than the discriminator, for instance, once in every five epochs.
\begin{equation}
    L_{D} = max_{w\in W} E_{x \sim P_r} [D_w(x)] + E_{z \sim Z} [D_w(G_\theta(z)] 
\end{equation}
\begin{equation}
    L_{G} = min_{\theta} - E_{z \sim Z} [D_w(G_\theta(z)] 
\end{equation}

\subsection{Experiments}
We use the CREMA-D dataset \cite{HouweiCao2014CCEM} for our experiments, which consists of 6 different emotion classes and 3413 audio clips of 12 different text sentences from 91 speakers (Appendix \ref{Dataset}). In order to study the model improvements, we employ two different approaches to present our data (Fig. \ref{data-representation}). In the first approach, we consider word length distribution plots for a pair of word indices in a phrase and compare how the word lengths are distributed for the generated data against the real dataset. Since this method only provides insight into a subset of word length space, another approach is to summarize the dataset using mean and variance to compare both distributions. However, the resulting plot is a concentrated graph which is hard to visually understand. So we consider word length distribution plots for all our experiments and compare models trained for 3 emotion classes i.e \textit{Happy}, \textit{Angry}, and \textit{Neutral} for 1000 epochs each. We also consider mean square error to understand how far generated data is from the mean of the real population, as presented in Table \ref{MSE_exp}.
\subsubsection{Conditioning on speaker}
As shown in Fig. \ref{emotion-box}, there is a significant variance in speech rates of different speakers, and this information is missed when the model is conditioned on only emotion. This was expected given the results of our exploration in Appendix \ref{stats}. This results in low-quality generated word lengths, as seen in distribution plot Fig. \ref{data-plots}a. To incorporate speaker information, we introduce a speaker encoder that takes in a 1-hot encoded vector of speaker identity and produces a latent code,  which is then concatenated along with the outputs from the text and emotion encoders. Hence, the model is now conditioned on emotion and speaker.
\subsubsection{Using reconstruction loss}
In addition to the Wasserstein loss defined in Section \ref{loss_func}, the model is intermittently trained on reconstruction loss between the generated word lengths and the ground truth. This helps the model to converge faster and generates a better data distribution, as seen in Fig. \ref{training-plots}a and Fig \ref{data-plots}e,c respectively.
\subsubsection{Using dynamic length input}
\begin{figure}[ht]
      \centering      \includegraphics[width=0.9\linewidth,height=0.22\linewidth]{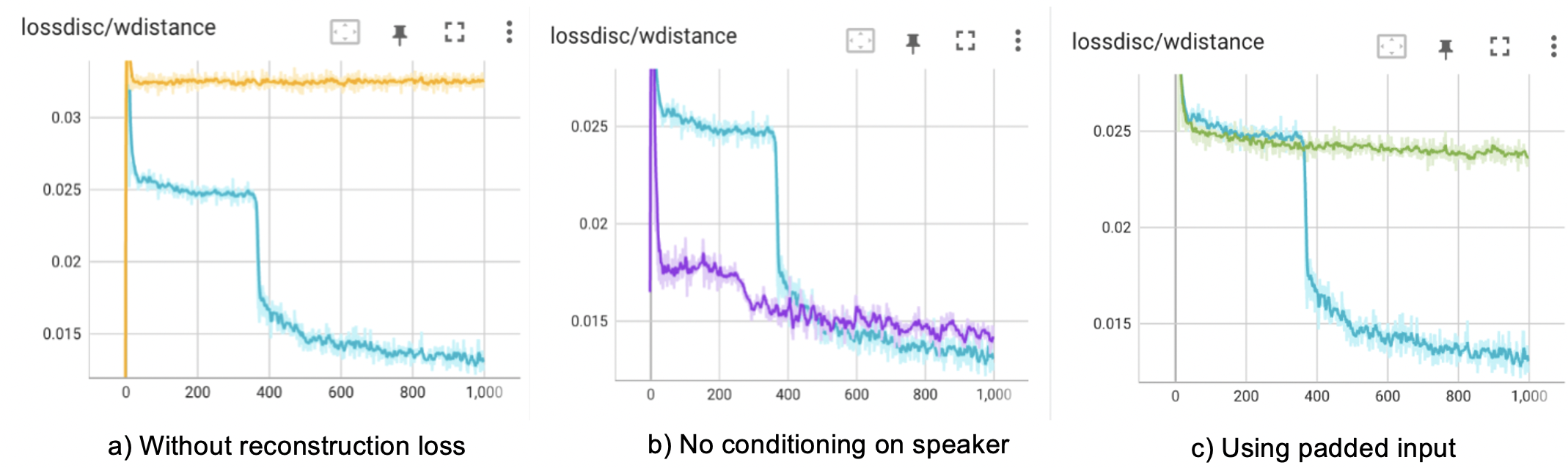}
      \caption{Discriminator loss for different settings, as compared to the final model (cyan line)}
      \label{training-plots}
\end{figure}

The dataset comprises 12 sentences of varying lengths. A usual approach to handle the varying length input is to pad the sequence with 0s or a small constant. Using the padded input, the model does not converge and generates undesired outputs (Figure \ref{data-plots}b), which can be reasoned with the nature of our target data. Since we are dealing with relative word lengths to neutral speech, most of the target data values are 0 or very small, which presents sparsity. With additional padding of 0s, the target data becomes more sparse and as misleading. We use gradient accumulation over a sequence of inputs to support the batch training of the model which provides a better convergence, as shown in Fig. \ref{training-plots}c.
\subsubsection{Using POS tags as input}
\begin{figure}[ht]
      \centering      \includegraphics[width=1\linewidth,height=0.38\linewidth]{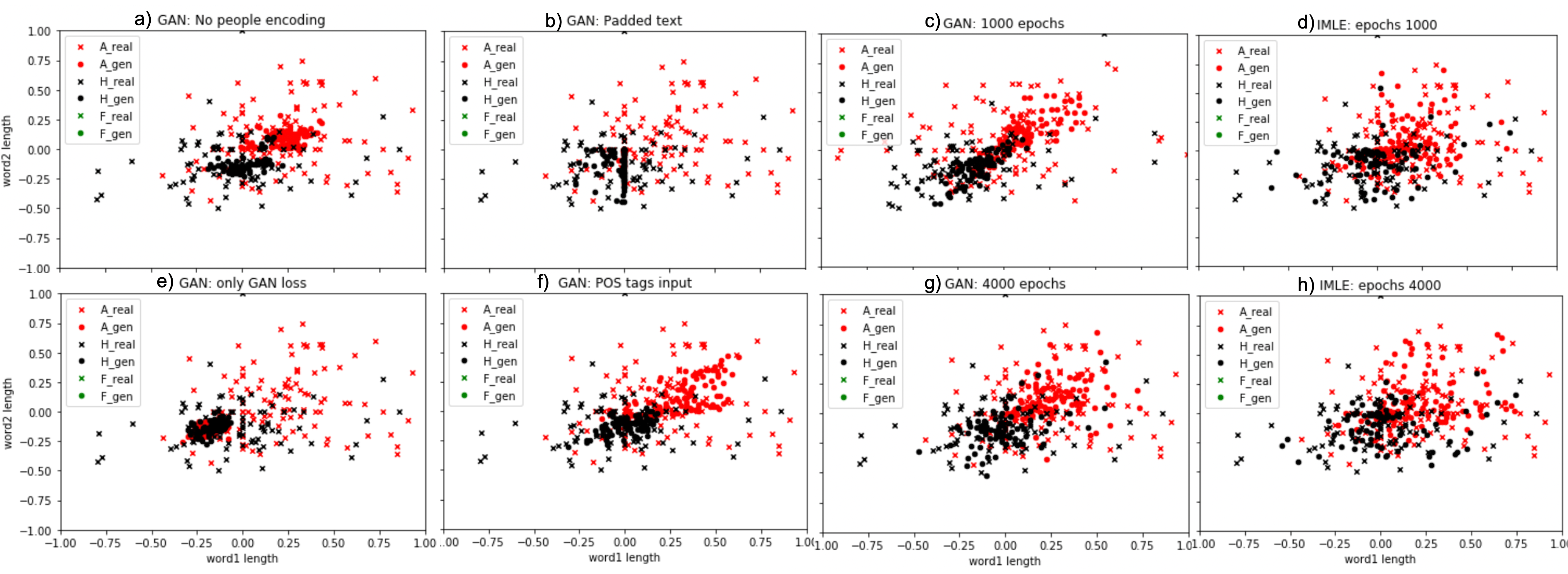}
      \caption{Generated word length distribution against empirical distribution for different experiments using GAN and IMLE model. The dots and the crosses represent generated and real data respectively.  }
      \label{data-plots}
\end{figure}

The POS tags capture the structure of a sentence which implies that any model trained on POS tags for actual text should be capable to generalize better on large datasets. Because of this we initially used POS tags as input. Although the resulting model converges, the generated dataset shows characteristics of mode collapse and does not capture the spread of real distribution (Fig. \ref{data-plots}f).


\begin{minipage}[c]{0.65\linewidth}
\centering
\begin{tabular}{|c|c|}
    \hline
        \textbf{Settings} & \textbf{MSE} \\
        \hline
        GAN w/o speaker condition & 0.398 \\
         GAN w/o reconstruction loss & 0.750 \\
         GAN with padding & 0.478 \\
         \raggedright GAN with POS tags & 0.374 \\ \hline
     \end{tabular}
     \label{MSE_exp}
\end{minipage}
\begin{minipage}[c]{0.3\linewidth}
\centering
\begin{tabular}{|c|c|c|}
    \hline
        \textbf{Settings} & \textbf{epochs} & \textbf{MSE} \\
        \hline
        \raggedright GAN & 1000 & 0.365 \\
        \raggedright GAN & 4000 & \textbf{0.298} \\
        \raggedright IMLE & 1000  & 0.332 \\
        \raggedright IMLE & 4000 & 0.303 \\ \hline
    \end{tabular}
    \label{GAN_IMLE}
\end{minipage}

\subsection{Implicit maximum likelihood estimation (IMLE)}
\begin{figure}[H]
      \centering
      \includegraphics[width=0.9\linewidth,height=0.2\linewidth]{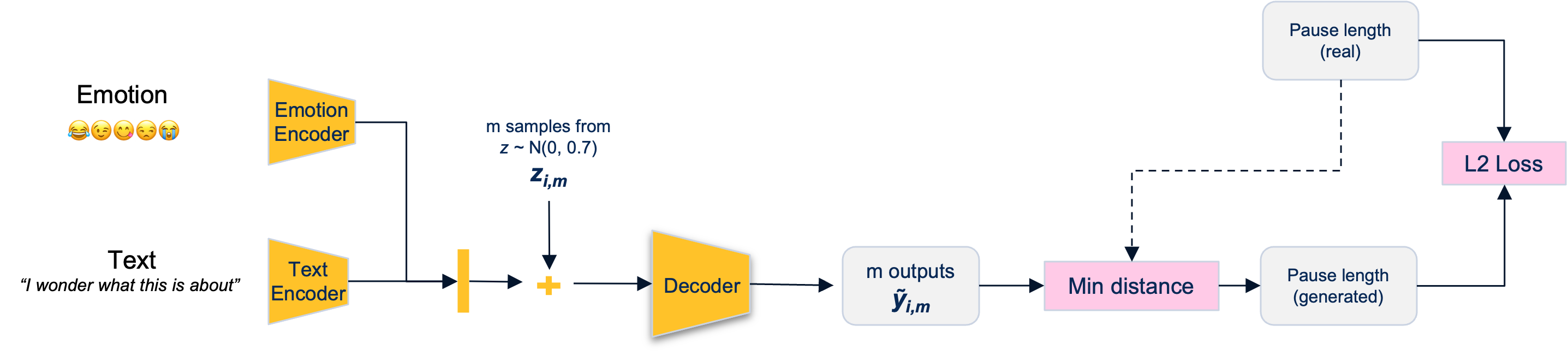}
      \caption{IMLE model pipeline}
      \label{imle-model}
\end{figure}

We extend our experiments by training an IMLE model to learn the data distribution and generate diverse word lengths. As presented in Fig. \ref{imle-model}, the IMLE model is similar to the GAN architecture  \ref{GAN-model}, with differences in noise injection and loss function. We use text and emotion encoders to get a generated latent code. Next, $m$ samples of noise from normal distribution $N(0, 0.7)$ (where 0.7 is the variance of the latent codes from pre-trained model) are added to the generated latent code independently and fed into the decoder to generate $m$ different outputs. Among these outputs, a sample with the minimum distance from a ground truth sample is chosen and euclidean distance or L2 loss is calculated for optimization. To help the model train faster, the encoders and decoder are pre-trained to predict the word length sequence (without using noise as input) and optimized using L2 loss between the predicted and ground truth sequence. 

Since the IMLE model learns to pull the generated samples closer to the real data, after 1000 epochs, the generated samples are more representative of true data when compared to GAN results after the same number of epochs (Fig. \ref{data-plots}c,d). After 4000 epochs, the generated data from the IMLE model still appears more diverse than the GAN results. As shown in Table \ref{GAN_IMLE}, the MSE values are slightly higher for IMLE-generated data which is potential because GANs learn a few modes of data very well while not considering for rest of the data. On the other hand, the IMLE model has tried to learn the entire distribution, introducing more noise for all the data points, which could improve by training the model further.

\section{Results}

\subsection{Objective Results}
To evaluate our model's performance we generated ground truth audio samples by directly feeding word lengths to Azure's TTS system\footnote{https://azure.microsoft.com/en-us/services/cognitive-services/text-to-speech/} using "en-US-JennyNeural" and conditioned on a given emotion. These word lengths were the average over all participants and intensities for each phrase. We then generated the baselines, which were the out-of-box Azure TTS for each emotion using the same SSML setup as the ground truth without word lengths. We once again used Gentle Aligner to extract word lengths from the baseline. 

Three evaluation methods were used, RMSE, Peasons correlation coefficient (PCC) and Dynamic time warping (DTW) with frame disturbance (FD). As our model is non-deterministic the PCC and RMSE are calculated as an average over all generated voices for all phrases. The results can be seen in table \ref{results}.

DTW allows us to assess how well-aligned the amplitudes are for two waveforms. The frame disturbance is the average sum of the squared difference in this alignment. Essentially this allows us to see how well-aligned the timing is for two signals of the same amplitude. The resulting DTW for the baseline as well as one of our generated phrases can be seen in Appendix \ref{DTW}.

\begin{table}
    \centering
    \begin{tabular}{|c|c|c|c|c|c|c|c|c|c|}
    \hline
    & \multicolumn{3}{|c|}{Angry} & \multicolumn{3}{|c|}{Happy} & \multicolumn{3}{|c|}{Neutral}\\ \hline
         \textbf{Model} & \textbf{RMSE} & \textbf{PCC} & \textbf{FD} &  \textbf{RMSE} & \textbf{PCC} & \textbf{FD} & \textbf{RMSE} & \textbf{PCC} & \textbf{FD}\\ \hline
         Baseline & 0.059 & 0.968 & 0.201 & 0.054 & 0.965 & 0.139 & 0.040 & 0.982 & n/a\\
         GAN & 0.096 & 0.969 & 0.403 & 0.060 & 0.961 & 0.009 & 0.000 & 1.000 & n/a \\ \hline
    \end{tabular}
    \caption{Objective results comparison between generated TTs and baseline TTS}
    \label{results}
\end{table}

\subsection{Subjective Results}
Due to the time and resource limitations of human participants we have not completed a thorough subjective evaluation. Ideally, a mean opinion score (MOS) and AB preference test would be conducted. However, the baseline, ground truth and generated text can be listened to in our supplemental material. As our model is deterministic, unlike the baseline, we generated 2 versions of each phrase for both angry and happy voices. 

\section{Conclusion}
In this work, we generated word lengths for speech relative to neutral emotion, conditioned on emotion and speaker. We experimented with GAN and IMLE models for generation where IMLE provides more diverse outputs while GANs provide lower mean square error from real datasets. The implementation can be further extended to more emotion classes and larger datasets with more diverse text in order to generalize the model. Overall our model performed well for happy and neutral speech but was lacking in its ability to generate accurate angry word lengths. We do, however, only have an objective assessment and the goal of emotion generation is perceptual. Given that objective results often do not reflect subjective performance \cite{TheisLucas2015Anot} a subjective assessment with several raters needs to be completed to be sure of the model's performance.




\bibliographystyle{plain}
\bibliography{references}

\appendix{}

\section{Statistical Tests}\label{stats}

\begin{figure}[H]
      \centering
      \includegraphics[scale=0.8]{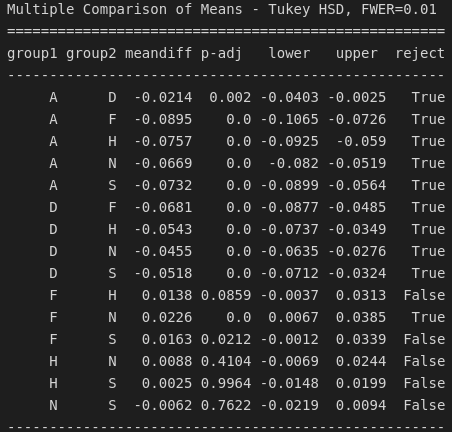}
      \caption{Table of Tukey test results with $\alpha = 0.01$, group differences between ROS for all emotions averaged over the speaker, intensity and phrase.}
      \label{emotion-tukey}
\end{figure}

\begin{figure}[H]
      \centering
      \includegraphics[scale=0.35]{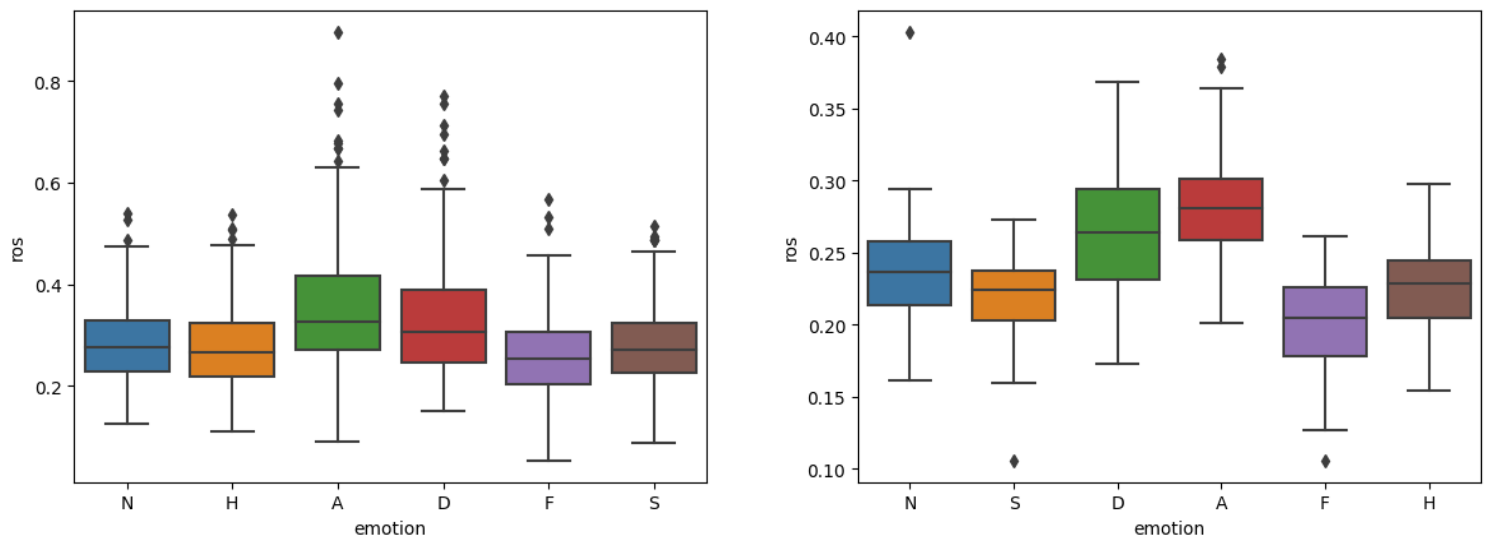}
      \caption{Left: Box plot of rate of speech for each emotion averaged over speaker, phrase, and intensity, Right: Box plot of rate of speech for each emotion for the phrase "We'll stop in a couple of minutes" averaged over speaker and intensity.}
      \label{emotion-box}
\end{figure}

In order to ensure that modelling speech rate over emotions would be feasible we conducted initial data exploration. We completed ANOVA's with our significance level set to $\alpha = 0.01$ to determine if the average rate of speech (ROS) was significantly different between emotions. Upon discovering significance in at least one of the emotion groups $F(5,3402)=95.89$, $p<0.001$. We completed a tukey test and found significance in all pairings except happiness with fear neutral and sadness, and fear and sadness; a full table can be seen in figure \ref{emotion-tukey}. A bar plot of the ROS for each emotion can been seen in fig. \ref{emotion-box} (left). In fig. \ref{emotion-box}(right) you can see the group differences are even more pronounced on a phrase level, however, we did not complete ANOVAs at this level or granularity.

We also wanted to explore possible confounding factors in the case that we need to make further modifications to the model at a later point. We completed ANOVA for group differences among the 91 speakers and found there to be a $F(90,3317)=4.34$, $p<0.001$ significance that at least one group is different. When completing tukey tests we found that approximately 4\% of speakers had a significantly different rate of speech averaged over all phrases and emotions. 

\begin{figure}[H]
\centering
\begin{minipage}{.4\textwidth}
      \centering
       \includegraphics[scale=0.35]{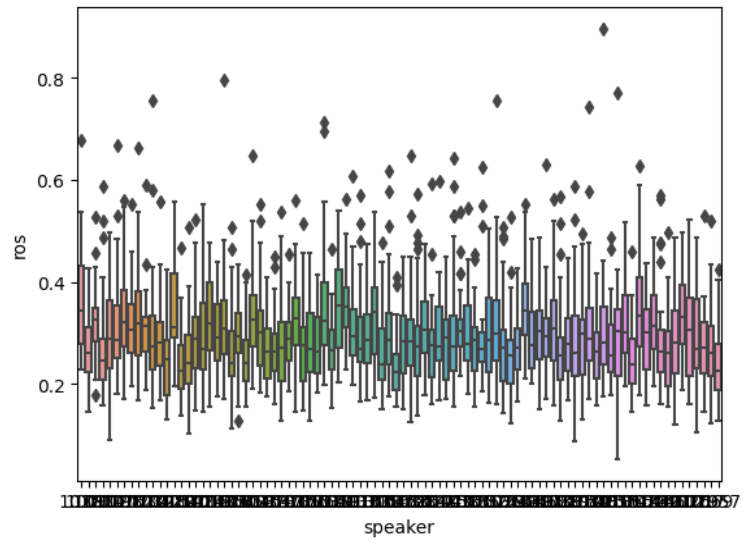}
      \caption{Box plot of rate of speech for each speaker averaged over intensity, phrase and emotion.}
      \label{speaker-box}
\end{minipage}%
\begin{minipage}{.1\linewidth}
..
\end{minipage}%
\begin{minipage}{.4\textwidth}
      \centering
      \includegraphics[scale=0.75]{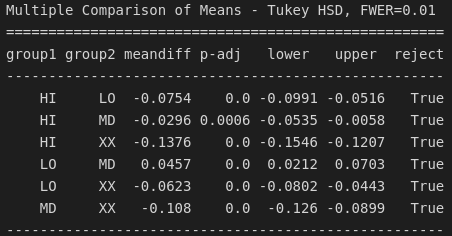}
      \caption{Table of Tukey test results with $\alpha = 0.01$, group differences between ROS for all intensities averaged over the speaker, emotion, and phrase.}
      \label{intensity-tukey}
\end{minipage}
\end{figure}

We also looked at group differences for intensity levels and found all intensities  to have significantly different ROS from one another, see fig. \ref{intensity-tukey}. It is unclear how these differences interact with one another and which contribute most to changes in speech rate. However, we keep in mind these possible confounding significant differences as we move forward in modelling.

\section{Dynamic time warping}\label{DTW}
DTW is used to find a time alignment of two signals of similar amplitude patterns. The chroma representation of the two waveforms are calculated then aligned into a cost matrix. The warping path is seen as the red line running diagonally thought the cost matrix. If perfectly aligned this will be a straight diagonal, any deviations are the results of misalignment. The root of the sum of squared difference between the $x$ and $y$ values of this line is the FD.

\begin{figure}[H]
\centering
\begin{minipage}{.4\textwidth}
      \centering
      \includegraphics[scale=0.35]{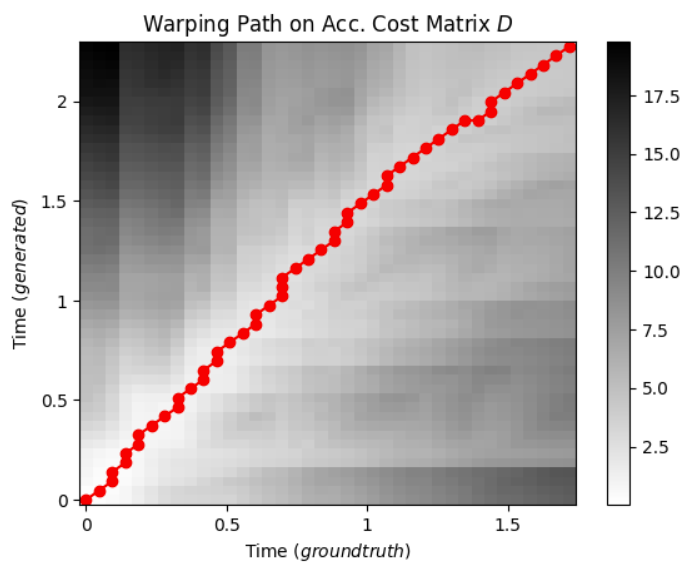}
      \caption{Dynamic time warping path for an angry phrase generated by the GAN compared to the ground truth.}
      \label{our-angry}
\end{minipage}%
\begin{minipage}{.1\linewidth}
..
\end{minipage}%
\begin{minipage}{.4\textwidth}
      \centering
      \includegraphics[scale=0.35]{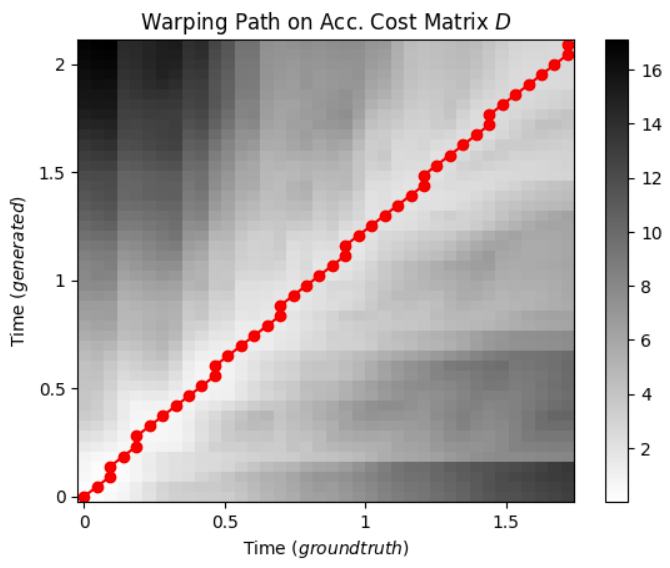}
      \caption{Dynamic time warping path for an angry phrase generated by the baseline compared to the ground truth.}
      \label{their-angry}
\end{minipage}
\end{figure}

\begin{figure}[H]
\centering
\begin{minipage}{.4\linewidth}
      \centering
      \includegraphics[scale=0.35]{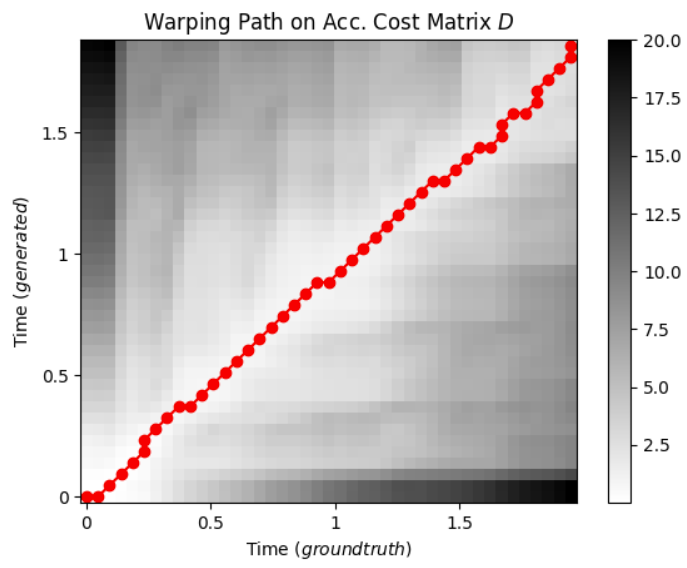}
      \caption{{Dynamic time warping path for a happy phrase generated by the GAN compared to the ground truth.}}
      \label{our-happy}
\end{minipage}%
\begin{minipage}{.1\linewidth}
..
\end{minipage}%
\begin{minipage}{.4\linewidth}
      \centering
      \includegraphics[scale=0.35]{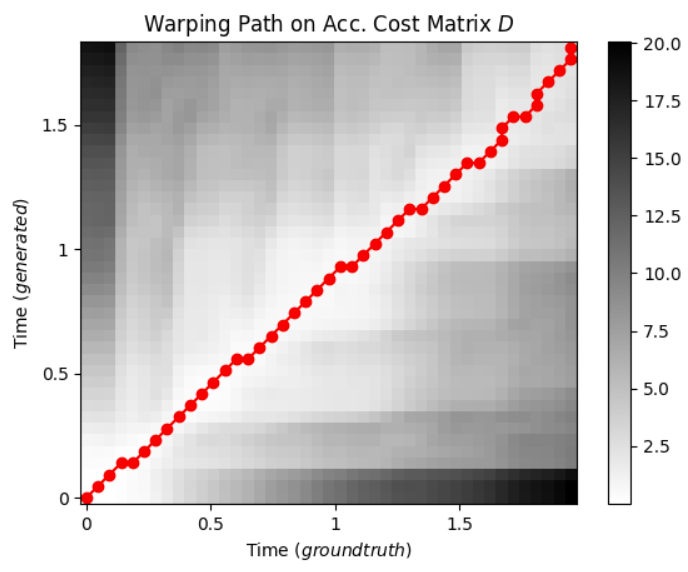}
      \caption{{Dynamic time warping path for a happy phrase generated by the baseline compared to the ground truth.}}
      \label{their-happy}
\end{minipage}
\end{figure}

DTW can also be displayed directly as an alignment of two waveforms. This has a less objective measure of comparison but allows to to see where the word lengths are and are not aligning well with the ground truth.

\begin{figure}[H]
\centering
\begin{minipage}{.43\textwidth}
      \centering
      \includegraphics[scale=0.32]{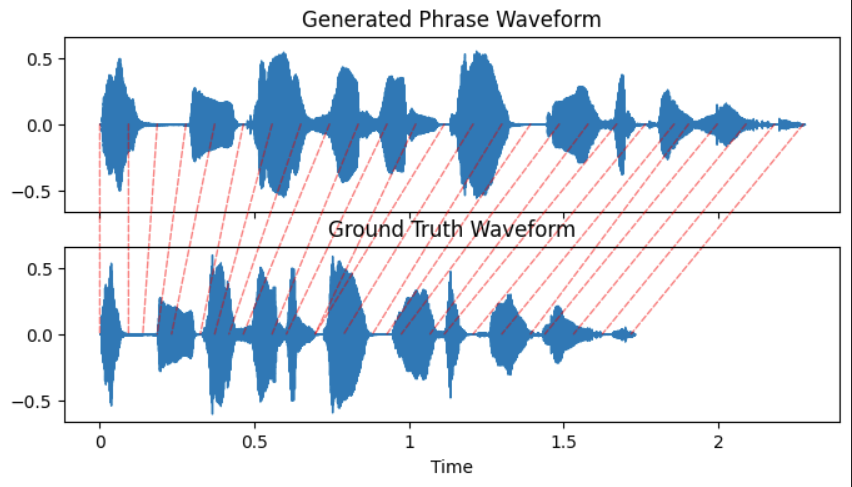}
      \caption{Dynamic time warping across waveforms for an angry phrase generated by the GAN compared to the ground truth.}
      \label{our-angry-wav}
\end{minipage}%
\begin{minipage}{.04\linewidth}
..
\end{minipage}%
\begin{minipage}{.43\textwidth}
      \centering
      \includegraphics[scale=0.32]{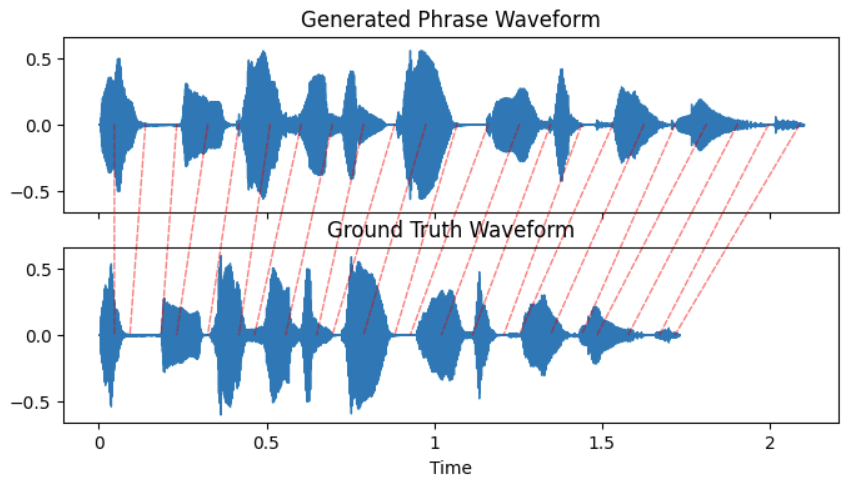}
      \caption{Dynamic time warping across waveforms for an angry phrase generated by the baseline compared to the ground truth.}
      \label{their-angry-wav}
\end{minipage}
\end{figure}

\begin{figure}[H]
\centering
\begin{minipage}{.43\textwidth}
      \centering
      \includegraphics[scale=0.32]{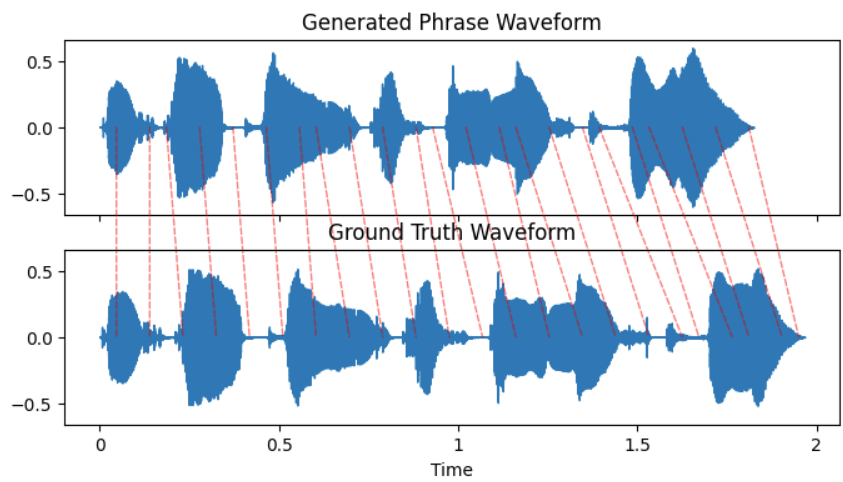}
      \caption{Dynamic time warping across waveforms for a happy phrase generated by the GAN compared to the ground truth.}
      \label{our-happy-wav}
\end{minipage}%
\begin{minipage}{.04\linewidth}
..
\end{minipage}%
\begin{minipage}{.43\textwidth}
      \centering
      \includegraphics[scale=0.32]{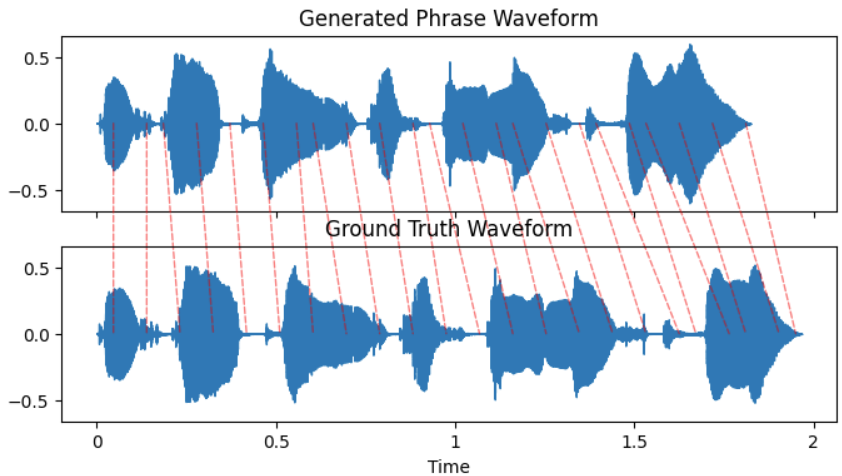}
      \caption{Dynamic time warping across waveforms for a happy phrase generated by the baseline compared to the ground truth.}
      \label{their-happy-wav}
\end{minipage}
\end{figure}

\section{CREMA-D text}\label{crema_sentences}
The sentences in CREMA-D dataset are as follows:
\begin{enumerate}
    \item I would like a new alarm clock
    \item I think I have a doctor's appointment
    \item Don't forget a jacket
    \item I think I've seen this before
    \item The surface is slick
    \item We'll stop in a couple of minutes
    \item It's eleven o'clock
    \item That is exactly what happened
    \item I'm on my way to the meeting
    \item I wonder what this is about
    \item The airplane is almost full
    \item Maybe tomorrow it will be cold
\end{enumerate}

\section{Additional content} \label{Figures}
\begin{figure}[H]
      \centering
      \includegraphics[width=0.7\linewidth,height=0.25\linewidth]{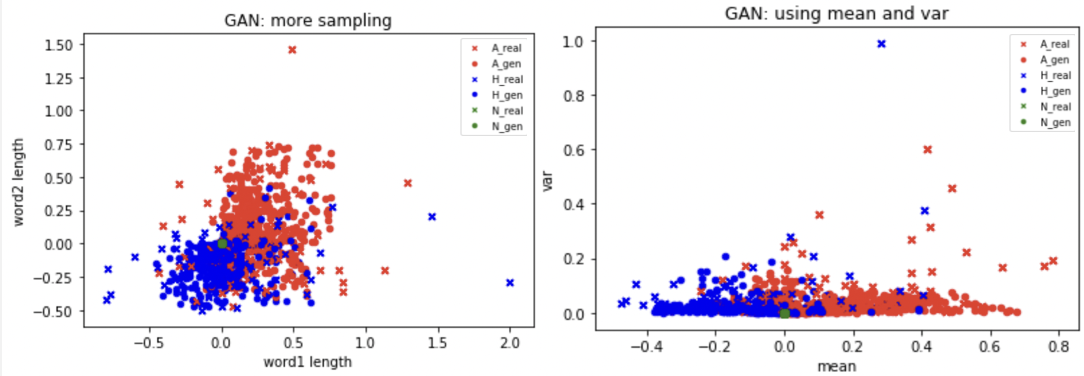}
      \caption{Proxy of data distribution a) pair of word lengths b) mean and variance of the sequence}
      \label{data-representation}
\end{figure}

\begin{table}[H]
    \centering
    \begin{tabular}{|c|c|c|c|c|c|}
    \hline
         \textbf{} & \textbf{Fear} & \textbf{Anger} & \textbf{Sadness} &  \textbf{Happiness} & \textbf{Disgust} \\ \hline
         \textbf{Speech rate} & very fast & slightly fast & slightly slow & fast or slow & very slow \\ 
         \textbf{Pitch} & very high & very high & slight low & high & very high \\ 
         \textbf{Pitch range} & very wide & very wide & slight narrow & very wide & slightly wide \\ 
         \textbf{Pitch changes} & normal & abrupt on stressed syllables & downward  & smooth upward  & wide downward  \\ 
        \textbf{Voice quality} & irregular & breathy chest tone & resonant & breathy blaring & grumbled chest tone \\ 
        \textbf{Articulation} & precise & tense & slurring & normal & normal \\
         \textbf{Intensity} & normal & high & low & high & low \\ \hline
    \end{tabular}
    \caption{Summary of human vocal effects most commonly associated with the emotion indicated and measured relative to neutral speech \cite{emotiontable}.}
    \label{emotion-table}
\end{table}


\end{document}